\pdfoutput=1

\documentclass[11pt]{article}

\usepackage[]{emnlp2021}

\usepackage{times}
\usepackage{latexsym}
\usepackage[normalem]{ulem}
\useunder{\uline}{\ul}{}
\usepackage[T1]{fontenc}

\usepackage[utf8]{inputenc}

\usepackage{microtype}

\usepackage[utf8]{inputenc} 
\usepackage[T1]{fontenc}    
\usepackage{hyperref}       
\usepackage{url}            
\usepackage{booktabs}       
\usepackage{amsfonts}       
\usepackage{nicefrac}       
\usepackage{microtype}      
\usepackage{xcolor}         
\usepackage{times}
\usepackage{latexsym}
\usepackage{graphicx}
\usepackage{multirow}
\usepackage{caption}

\title{Temporal Word Meaning Disambiguation using TimeLMs}

\newcommand{\repeatthanks}{\textsuperscript{\thefootnote} }

\author{Mihir Godbole\thanks{~~Equal Contribution} \hspace{0.5pt} \and Parth Dandavate\repeatthanks \and Aditya Kane\repeatthanks
  \\
  Pune Institute of Computer Technology, Pune\\
  \texttt{\{mihirgod11, dandavateparth, adityakane1\}@gmail.com}}

\begin{document}
\maketitle
\begin{abstract}
    Meaning of words constantly change given the events in modern civilization. Large Language Models use word embeddings, which are often static and thus cannot cope with this semantic change. Thus,it is important to resolve ambiguity in word meanings. This paper is an effort in this direction, where we explore methods for word sense disambiguation for the EvoNLP shared task. We conduct rigorous ablations for two solutions to this problem. We see that an approach using time-aware language models helps this task. Furthermore, we explore possible future directions to this problem.
\end{abstract}

\section{Introduction}
A change in the meaning of a word in varying semantic contents is a challenge for various NLP tasks such as text and sentence classification, question answering and sentence prediction. Recent developments in large language models (LLMs) like ELMo  \cite{peters-etal-2018-deep}, BERT \cite{devlin-etal-2019-bert} and GPT  \cite{brown2020language} have revolutionised the field of NLP with context dependent word embeddings. These models have been trained on a large corpus of unlabelled text. While these models take in consideration the semantics of the text, it is limited to the corpus it was trained on. This introduces a new challenge of the shift in the meaning of a word across the temporal axis.

Word Sense Disambiguation \cite{huang-etal-2019-glossbert} is the process of identifying the meaning of a word from multiple possible meanings in varying contexts. This task can be further extended as a polysemy resolution task to classify the meaning of words in different contexts. Our system performs a similar task while classifying two texts with  a common word with the same or different meaning. Specifically, the premise of our system is to classify tweets from two different time periods with a common word. The variation in the meaning of a word is caused by two factors, the context of the word in the form of a tweet or a change in the usage and hence in the meaning of a word because of the shift along the time axis. Historically it has been observed that the meanings of some words have been altered over time. For example, the word "fathom" originally meant "to encircle with one’s arms” and now is defined as “to understand after much thought”. The ever expanding nature of the internet and social media have led to rapid evolution of words, with the meanings of words changing and new words getting csoined. This means that the corpus of data used for training a LLM will keep changing over time. Hence, the pretrained models for existing LLMs like BERT, RoBERTa cannot be used to compare word embeddings for a word from two different time periods. This shared task \cite{loureiro-etal-2022-tempowic} focusses precisely on this problem statement.

To address this problem, we propose a system comprising of 
TimeLMs \cite{loureiro-etal-2022-timelms} to incorporate the time aspect of the data. TimeLMs are language models that are trained using data up to a certain time instance. In this case they are trained on tweets gathered by the end of a year. Therefore there exists a unique TimeLM model for each year which takes into account the time aspect of data. The dataset used for testing our system consists of tweets from the years 2019, 2020 and 2021. Tweets from two different time periods containing a common word are paired in this dataset and labelled to indicate similarity or dissimilarity in the meanings of that word in the two tweets. The TimeLMs used in our system are Roberta models trained on tweets upto the specific year. This enables our system to get an accurate representations of the words based on their use upto that time period. The embeddings are then compared based on a similarity metric to classify the tweets using a preset threshold value for similarity. 

This paper is organised as follows. We analyse existing research and methods in Section (2). We give a overview of the dataset used for our system in Section (3). We provide a overview of our system implementation in Section (4). We also compare the results of our experiments in developing this system in Section (5). We discuss the possible improvements and scope of this system in Section (6). 

\begin{figure*}[ht]
    \centering
    \includegraphics[width=\textwidth]{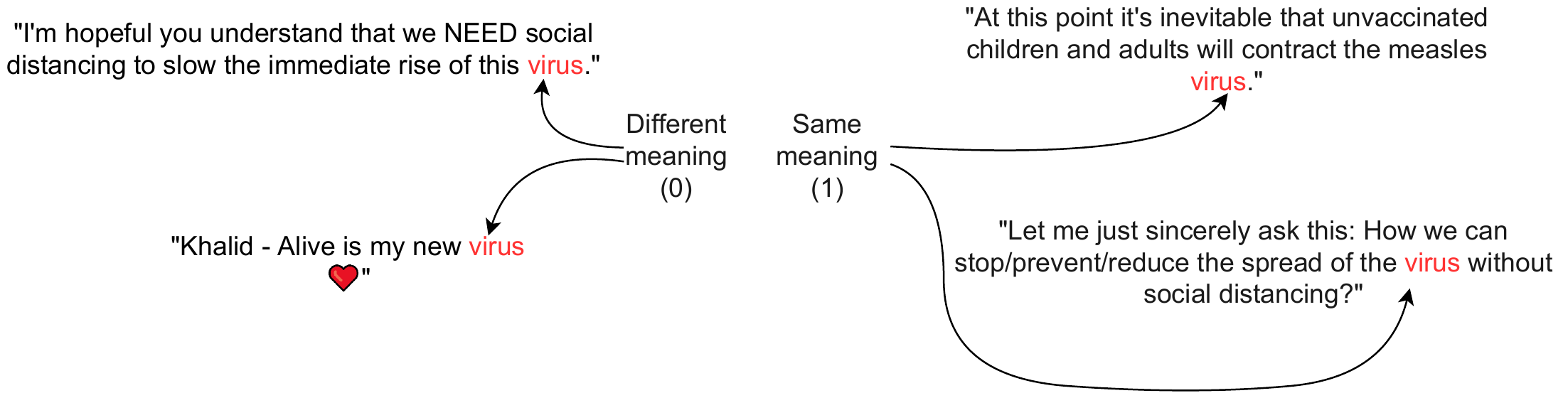}
    \caption{Examples from the datset.}
    \label{fig:data_example}
\end{figure*}

\section{Related work}

In Natural Language Processing, the meaning of words is denoted by a vector, commonly known as word embedding. Works like GloVe \cite{pennington2014glove} and Word2Vec \cite{mikolov2013word2vec} were one of the first ones to represent a word using vectors. However, the embeddings thus generated were context-agnostic, meaning their meaning was fixed and were not dependent on the context. 

With the dawn of modern text encoders \cite{vaswani2017attention, devlin-etal-2019-bert}, context dependent embeddings can be easily calculated. Works like \citeauthor{wic, xlwic} aim to have manually annotated datasets containing pairs of sentences having same or different meaning, and labelling them as such. To solve this task, several methods have been developed. Works like \citet{levine-etal-2020-sensebert, peters-etal-2019-knowledge} try to impart context based knowledge into the embeddings by using WordNet \cite{wordnet} attributes. The models are trained in a self-supervised fashion with entity linking. Another approach is to use word-level embeddings. \citet{loureiro-jorge-2019-language} use this approach, combining it with a $k$-NN ($k$ Nearest Neighbours) method to disambiguate the word embeddings. Note that transformers can also be used for this purpose, since the output features from the transformers can be interpreted as word embeddings. \citet{Loureiro2022LMMSRT} studies model layers to understand the effect of attention-based architectures in word sense disambiguation task. Elmo \cite{peters-etal-2018-deep} is one of many available architectures in this direction. Lastly, work has been done to incorporate the semantic space knowledge into the embeddings \cite{colla-etal-2020-lesslex}, also known as sense-based disambiguation.

Given this, little work has been done on word meaning disambiguation in a temporal setting. This means that the information about the time of text utterance is also provided along with the sentence itself. This paper tries to provide a solution to this problem - word meaning disambiguation when temporal information is available.


\section{Dataset description}
\label{sec:dataset}


The dataset consists of 1428 training samples and 396 validation samples. The final scores were calculated on a set of 10,000 unseen test samples. In every training sample, we were provided with two sentences and the word whose semantic meaning was to be compared. Some metadata like tokens and start and end of word was also included in every sample.In the training dataset, out of the 1428 samples, 650 examples had the words in two sentences having same meaning, whereas 778 samples had the words in two sentences having different meaning. Note that since this dataset is relatively balanced, and hence does not need any additional preprocessing to balance the data distribution. However, it is important to note that the target words in the training and testing dataset constitute two different sets, and hence the problem should be solved in a way that is target word agnostic.

 An illustration of the data is shown in Figure \ref{fig:data_example}. The left part shows an example where the meaning of the target word "virus" is different in both tweets. Specifically, in the top left tweet it indicates to the disease-causing organism whereas the bottom left tweet indicates to a thing that the person likes. In the right part, both instances of the target word mean the same, denoting disease-causing organism.

The dataset also provides the month and year when the tweet was written. This provides us the temporal information, which can be useful for the semantic evaluation of the words in the given context. Our approach aims at using this semantic information in a way that a language model relevant to the tweet is used to get the semantic features of the tweet. 

\section{Methodology}
\begin{figure*}
\centering
\begin{minipage}{.5\textwidth}
  \centering
  \includegraphics[width=.6\textwidth]{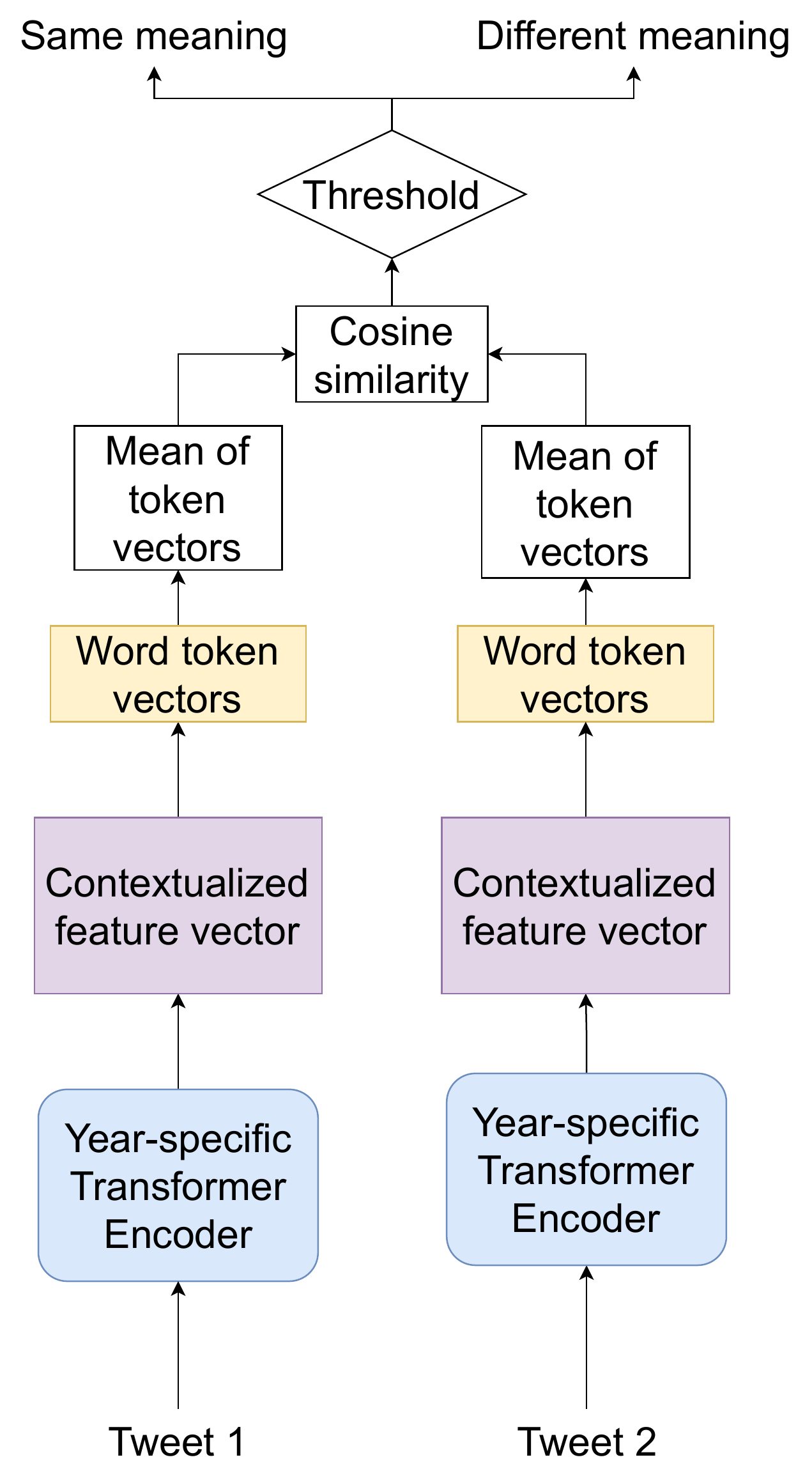}
  \captionof{figure}{TimeLMs aided word sense disambiguation}
  \label{fig:timelms_method}
\end{minipage}%
\begin{minipage}{.5\textwidth}
  \centering
  \includegraphics[width=.8\textwidth]{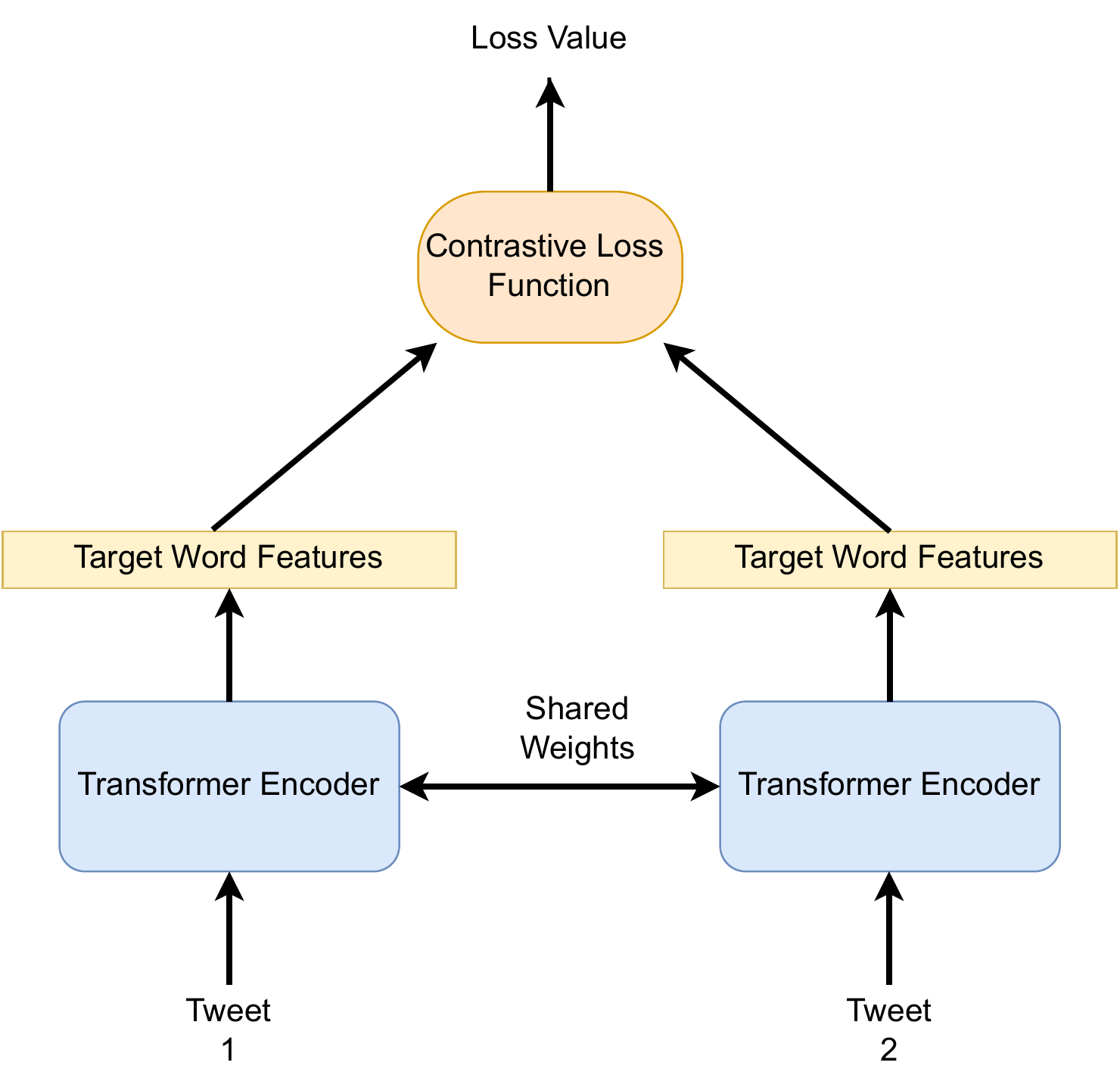}
  \captionof{figure}{Contrastive feature based classification}
  \label{fig:test2}
\end{minipage}
\end{figure*}
\subsection{TimeLMs aided word sense disambiguation}

As mentioned in Section \ref{sec:dataset}, the date of posting of the tweet is provided as a data attribute. In this approach, we used this information to choose the transformer model to extract target features. We use TimeLMs \cite{loureiro-etal-2022-timelms} for this purpose. We observe this performs better compared to using a single model. Our method is illustrated in Figure \ref{fig:timelms_method}.

Specifically, we observe that the tweets in the input data are posted in the years 2019 and 2020 only. Thus, we use the variants of TimeLMs trained on Twitter data collected until December 2019 and 2020 for respectively dated tweets. In this way, we can encapsulate the difference in semantic representations of sentences across time. 

After extracting the contextualized sentence features from the respective models, we extract the target word features. We hereby get two word feature vectors, one corresponding to each tweet. Note that since one word may be split into multiple tokens, we use the mean of these token-wise features for out computation. Note that the feature vectors for a tweet is the mean of the last four layers of the language models concatenated to the pooled ($[CLS]$ token) output. These two feature vectors are then compared with each other using cosine similarity. If this cosine similarity is high, the meaning of the target word in two sentences is the same, alternatively if the cosine similarity is low then the meaning of the target word in the two sentences is different. 

Since this approach does not actually train the parameters of the model, we use the training dataset to calculate the thresholds. Specifically, we iterate over potential thresholds between 0 and 1 with a step of 0.001. We then rank the thresholds based on their F1 scores. The best performing threshold is then used for generating the final predictions. The threshold for our best performing model (TimeLMs) was $0.917$. We use five models for our ablations: ELECTRA (small) \cite{electra}, ALBERT (base) \cite{albert}, BERT (base, uncased) \cite{devlin-etal-2019-bert}, RoBERTa (base) \cite{zhuang-etal-2021-robustly}, TimeLMs \cite{loureiro-etal-2022-timelms}.

\subsection{Contrastive feature based classification}
The task essentially being identifying whether the usage of word is similar or not we thought of training the language models in a Siamese setting. Siamese networks involves two similar encoder networks with the same weights and a classification system, which determines the similarity based on the distance between encoded features and a threshold. As mentioned in the previous sub section we are extracting the target word features using transformer models which will be the encoders. If the meaning of the word in both the sentences is same then the target word features given by the transformer model should be similar. 

We trained the model using a simple contrastive loss involving euclidean distance between the target word features. We used the same models as mentions in the previous sections, except for the TimeLMs. For determining the threshold for the classification process we iterated through a range of 0 to 4, with a step of 0.01, while testing on the validation data. The threshold was determined for the euclidean distance between the word embeddings obtained from the model. The threshold for our best performing
model (TimeLMs) was 1.148.

\subsection{Implementation details}

We use the HuggingFace library \cite{wolf-etal-2020-transformers} for our experiments. For the cosine similarity experiments, we find a threshold of 0.917 for our best performing solution. We use a batch size of 64. Here the threshold was selected for the cosine distance between the two word embeddings.

For the contrastive method experiments, we find a threshold of 1.148 for our best performing solution (RoBERTa). We used a batch size of 8. Here the threshold was selected for the euclidean distance between the two word embeddings.

In both cases, the inference distance value (cosine or euclidean) below the threshold indicated similar meaning for the two words, and the the inference distance value above the threshold indicated different meaning for the two words.

\section{Results}


\begin{table}[]
\begin{tabular}{|l|r|r|r|}
\hline
\multicolumn{1}{|c|}{\textbf{Model}} & \multicolumn{1}{c|}{\textbf{\begin{tabular}[c]{@{}c@{}}Val \\ F1-score\end{tabular}}} & \multicolumn{1}{c|}{\textbf{\begin{tabular}[c]{@{}c@{}}Val \\ Accuracy\end{tabular}}} & \multicolumn{1}{c|}{\textbf{\begin{tabular}[c]{@{}c@{}}Test \\ F1-score\end{tabular}}} \\ \hline
Electra                              & 61.00                                                                                    & 54.78                                                                                 & 38.77                                                                                  \\ \hline
RoBERTa                              & 60.00                                                                                    & 56.51                                                                                 & 38.96                                                                                  \\ \hline
BERT                                 & 60.80                                                                                  & 56.77                                                                                 & 38.77                                                                                  \\ \hline
Albert                               & 60.73                                                                                 & 56.77                                                                                 & 39.00                                                                                  \\ \hline
TimeLMs                              & \textbf{61}                                                                           & \textbf{61.71}                                                                        & \textbf{57.94}                                                                         \\ \hline
\end{tabular}
\caption{Results of Similarity Method}
\label{tab:cosine}
\end{table}
\begin{table}[]
\begin{tabular}{|l|r|r|r|}
\hline
\multicolumn{1}{|c|}{\textbf{Model}} & \multicolumn{1}{c|}{\textbf{\begin{tabular}[c]{@{}c@{}}Val \\ F1-score\end{tabular}}} & \multicolumn{1}{c|}{\textbf{\begin{tabular}[c]{@{}c@{}}Val \\ Accuracy\end{tabular}}} & \multicolumn{1}{c|}{\textbf{\begin{tabular}[c]{@{}c@{}}Test \\ F1-score\end{tabular}}} \\ \hline
Electra                              & \textbf{66.67}                                                                        & \textbf{75}                                                                           & 46.15                                                                                \\ \hline
RoBERTa                              & 60.8                                                                                  & 44.01                                                                                 & \textbf{48.97}                                                                       \\ \hline
BERT                                 & 65.44                                                                                 & 48.98                                                                                 & 44.34                                                                                \\ \hline
Albert                               & 66.6                                                                                  & 66.6                                                                                  & 43.75                                                                                \\ \hline
\end{tabular}
\caption{Results of Contrastive Method}
\label{tab:siamese}
\end{table}

We hereby present the results of both of our methods. We report several interesting observations based on the results.

Our results based on our cosine similarity are shown in Table \ref{tab:cosine} and our results based on the contrastive method are shown in Table \ref{tab:siamese}. 

\begin{enumerate}
    \item \textbf{TimeLMs based method performs the best: } We observe that the TimeLMs based method performs the best. We speculate this is because of the time-aware nature of the models. Some
    words, for example "lockdown" have significantly different meaning before and after the pandemix. Thus, models pretrained on the specific data results in better performance.
    \item \textbf{BERT and AlBERT have similar performance: } We see that BERT and Albert have very similar Accuracy and Macro-F1. We hypothesize that this is because of the similarity in their pretraining objectives. Albert is a model aimed to mimic the capabilities of BERT, but with lower number of parameters. Thus, it makes sense that these models have very similar validation metrics.

    \item  \textbf{Electra has a better language representation: } As seen on state of art benchmarks like GLUE and SQuAD Electra is outperforming RoBERTa, ALBERT. Electra has achieved better  F1-score and accuracy compared to both.

 \end{enumerate}


\section{Conclusion}
In this paper, we explore two solutions to the word sense disambiguation problem within the scope of EvoNLP shared task. We report a maximum testing F1-score of 57.94\% with TimeLMs.  We foresee several research directions for this work. One line of work can be explore robustness of the contrastive models. The threshold search technique for this method can be explored in greater detail.

\bibliography{anthology,custom}

\begin{thebibliography}{21}
\expandafter\ifx\csname natexlab\endcsname\relax\def\natexlab#1{#1}\fi

\bibitem[{Brown et~al.(2020)Brown, Mann, Ryder, Subbiah, Kaplan, Dhariwal,
  Neelakantan, Shyam, Sastry, Askell et~al.}]{brown2020language}
Tom~B Brown, Benjamin Mann, Nick Ryder, Melanie Subbiah, Jared Kaplan, Prafulla
  Dhariwal, Arvind Neelakantan, Pranav Shyam, Girish Sastry, Amanda Askell,
  et~al. 2020.
\newblock Language models are few-shot learners.
\newblock \emph{arXiv preprint arXiv:2005.14165}.

\bibitem[{Clark et~al.(2020)Clark, Luong, Le, and Manning}]{electra}
Kevin Clark, Minh{-}Thang Luong, Quoc~V. Le, and Christopher~D. Manning. 2020.
\newblock \href {https://openreview.net/forum?id=r1xMH1BtvB} {{ELECTRA:}
  pre-training text encoders as discriminators rather than generators}.
\newblock In \emph{8th International Conference on Learning Representations,
  {ICLR} 2020, Addis Ababa, Ethiopia, April 26-30, 2020}. OpenReview.net.

\bibitem[{Colla et~al.(2020)Colla, Mensa, and
  Radicioni}]{colla-etal-2020-lesslex}
Davide Colla, Enrico Mensa, and Daniele~P. Radicioni. 2020.
\newblock \href {https://doi.org/10.1162/coli_a_00375} {{L}ess{L}ex: Linking
  multilingual embeddings to {S}en{S}e representations of {LEX}ical items}.
\newblock \emph{Computational Linguistics}, 46(2):289--333.

\bibitem[{Devlin et~al.(2019)Devlin, Chang, Lee, and
  Toutanova}]{devlin-etal-2019-bert}
Jacob Devlin, Ming-Wei Chang, Kenton Lee, and Kristina Toutanova. 2019.
\newblock \href {https://doi.org/10.18653/v1/N19-1423} {{BERT}: Pre-training of
  deep bidirectional transformers for language understanding}.
\newblock In \emph{Proceedings of the 2019 Conference of the North {A}merican
  Chapter of the Association for Computational Linguistics: Human Language
  Technologies, Volume 1 (Long and Short Papers)}, pages 4171--4186,
  Minneapolis, Minnesota. Association for Computational Linguistics.

\bibitem[{Huang et~al.(2019)Huang, Sun, Qiu, and
  Huang}]{huang-etal-2019-glossbert}
Luyao Huang, Chi Sun, Xipeng Qiu, and Xuanjing Huang. 2019.
\newblock \href {https://doi.org/10.18653/v1/D19-1355} {{G}loss{BERT}: {BERT}
  for word sense disambiguation with gloss knowledge}.
\newblock In \emph{Proceedings of the 2019 Conference on Empirical Methods in
  Natural Language Processing and the 9th International Joint Conference on
  Natural Language Processing (EMNLP-IJCNLP)}, pages 3509--3514, Hong Kong,
  China. Association for Computational Linguistics.

\bibitem[{Lan et~al.(2019)Lan, Chen, Goodman, Gimpel, Sharma, and
  Soricut}]{albert}
Zhenzhong Lan, Mingda Chen, Sebastian Goodman, Kevin Gimpel, Piyush Sharma, and
  Radu Soricut. 2019.
\newblock \href {https://doi.org/10.48550/ARXIV.1909.11942} {Albert: A lite
  bert for self-supervised learning of language representations}.

\bibitem[{Levine et~al.(2020)Levine, Lenz, Dagan, Ram, Padnos, Sharir,
  Shalev-Shwartz, Shashua, and Shoham}]{levine-etal-2020-sensebert}
Yoav Levine, Barak Lenz, Or~Dagan, Ori Ram, Dan Padnos, Or~Sharir, Shai
  Shalev-Shwartz, Amnon Shashua, and Yoav Shoham. 2020.
\newblock \href {https://doi.org/10.18653/v1/2020.acl-main.423} {{S}ense{BERT}:
  Driving some sense into {BERT}}.
\newblock In \emph{Proceedings of the 58th Annual Meeting of the Association
  for Computational Linguistics}, pages 4656--4667, Online. Association for
  Computational Linguistics.

\bibitem[{Loureiro et~al.(2022{\natexlab{a}})Loureiro, Barbieri, Neves,
  Espinosa~Anke, and Camacho-collados}]{loureiro-etal-2022-timelms}
Daniel Loureiro, Francesco Barbieri, Leonardo Neves, Luis Espinosa~Anke, and
  Jose Camacho-collados. 2022{\natexlab{a}}.
\newblock \href {https://doi.org/10.18653/v1/2022.acl-demo.25} {{T}ime{LM}s:
  Diachronic language models from {T}witter}.
\newblock In \emph{Proceedings of the 60th Annual Meeting of the Association
  for Computational Linguistics: System Demonstrations}, pages 251--260,
  Dublin, Ireland. Association for Computational Linguistics.

\bibitem[{Loureiro et~al.(2022{\natexlab{b}})Loureiro, D{'}Souza, Muhajab,
  White, Wong, Espinosa-Anke, Neves, Barbieri, and
  Camacho-Collados}]{loureiro-etal-2022-tempowic}
Daniel Loureiro, Aminette D{'}Souza, Areej~Nasser Muhajab, Isabella~A. White,
  Gabriel Wong, Luis Espinosa-Anke, Leonardo Neves, Francesco Barbieri, and
  Jose Camacho-Collados. 2022{\natexlab{b}}.
\newblock \href {https://aclanthology.org/2022.coling-1.296} {{T}empo{W}i{C}:
  An evaluation benchmark for detecting meaning shift in social media}.
\newblock In \emph{Proceedings of the 29th International Conference on
  Computational Linguistics}, pages 3353--3359, Gyeongju, Republic of Korea.
  International Committee on Computational Linguistics.

\bibitem[{Loureiro and Jorge(2019)}]{loureiro-jorge-2019-language}
Daniel Loureiro and Al{\'\i}pio Jorge. 2019.
\newblock \href {https://doi.org/10.18653/v1/P19-1569} {Language modelling
  makes sense: Propagating representations through {W}ord{N}et for
  full-coverage word sense disambiguation}.
\newblock In \emph{Proceedings of the 57th Annual Meeting of the Association
  for Computational Linguistics}, pages 5682--5691, Florence, Italy.
  Association for Computational Linguistics.

\bibitem[{Loureiro et~al.(2022{\natexlab{c}})Loureiro, Jorge, and
  Camacho-Collados}]{Loureiro2022LMMSRT}
Daniel Loureiro, Al'ipio~M'ario Jorge, and Jos{\'e} Camacho-Collados.
  2022{\natexlab{c}}.
\newblock Lmms reloaded: Transformer-based sense embeddings for disambiguation
  and beyond.
\newblock \emph{Artif. Intell.}, 305:103661.

\bibitem[{Mikolov et~al.(2013)Mikolov, Chen, Corrado, and
  Dean}]{mikolov2013word2vec}
Tomas Mikolov, Kai Chen, Greg Corrado, and Jeffrey Dean. 2013.
\newblock \href {https://doi.org/10.48550/ARXIV.1301.3781} {Efficient
  estimation of word representations in vector space}.

\bibitem[{Miller(1995)}]{wordnet}
George~A. Miller. 1995.
\newblock \href {https://doi.org/10.1145/219717.219748} {Wordnet: A lexical
  database for english}.
\newblock \emph{Commun. ACM}, 38(11):39–41.

\bibitem[{Pennington et~al.(2014)Pennington, Socher, and
  Manning}]{pennington2014glove}
Jeffrey Pennington, Richard Socher, and Christopher~D. Manning. 2014.
\newblock \href {http://www.aclweb.org/anthology/D14-1162} {Glove: Global
  vectors for word representation}.
\newblock In \emph{Empirical Methods in Natural Language Processing (EMNLP)},
  pages 1532--1543.

\bibitem[{Peters et~al.(2018)Peters, Neumann, Iyyer, Gardner, Clark, Lee, and
  Zettlemoyer}]{peters-etal-2018-deep}
Matthew~E. Peters, Mark Neumann, Mohit Iyyer, Matt Gardner, Christopher Clark,
  Kenton Lee, and Luke Zettlemoyer. 2018.
\newblock \href {https://doi.org/10.18653/v1/N18-1202} {Deep contextualized
  word representations}.
\newblock In \emph{Proceedings of the 2018 Conference of the North {A}merican
  Chapter of the Association for Computational Linguistics: Human Language
  Technologies, Volume 1 (Long Papers)}, pages 2227--2237, New Orleans,
  Louisiana. Association for Computational Linguistics.

\bibitem[{Peters et~al.(2019)Peters, Neumann, Logan, Schwartz, Joshi, Singh,
  and Smith}]{peters-etal-2019-knowledge}
Matthew~E. Peters, Mark Neumann, Robert Logan, Roy Schwartz, Vidur Joshi,
  Sameer Singh, and Noah~A. Smith. 2019.
\newblock \href {https://doi.org/10.18653/v1/D19-1005} {Knowledge enhanced
  contextual word representations}.
\newblock In \emph{Proceedings of the 2019 Conference on Empirical Methods in
  Natural Language Processing and the 9th International Joint Conference on
  Natural Language Processing (EMNLP-IJCNLP)}, pages 43--54, Hong Kong, China.
  Association for Computational Linguistics.

\bibitem[{Pilehvar and Camacho{-}Collados(2018)}]{wic}
Mohammad~Taher Pilehvar and Jos{\'{e}} Camacho{-}Collados. 2018.
\newblock Wic: 10, 000 example pairs for evaluating context-sensitive
  representations.
\newblock \emph{CoRR}, abs/1808.09121.

\bibitem[{Raganato et~al.(2020)Raganato, Pasini, Camacho{-}Collados, and
  Pilehvar}]{xlwic}
Alessandro Raganato, Tommaso Pasini, Jos{\'{e}} Camacho{-}Collados, and
  Mohammad~Taher Pilehvar. 2020.
\newblock Xl-wic: {A} multilingual benchmark for evaluating semantic
  contextualization.
\newblock \emph{CoRR}, abs/2010.06478.

\bibitem[{Vaswani et~al.(2017)Vaswani, Shazeer, Parmar, Uszkoreit, Jones,
  Gomez, Kaiser, and Polosukhin}]{vaswani2017attention}
Ashish Vaswani, Noam Shazeer, Niki Parmar, Jakob Uszkoreit, Llion Jones,
  Aidan~N Gomez, \L~ukasz Kaiser, and Illia Polosukhin. 2017.
\newblock \href
  {https://proceedings.neurips.cc/paper/2017/file/3f5ee243547dee91fbd053c1c4a845aa-Paper.pdf}
  {Attention is all you need}.
\newblock In \emph{Advances in Neural Information Processing Systems},
  volume~30. Curran Associates, Inc.

\bibitem[{Wolf et~al.(2020)Wolf, Debut, Sanh, Chaumond, Delangue, Moi, Cistac,
  Rault, Louf, Funtowicz, Davison, Shleifer, von Platen, Ma, Jernite, Plu, Xu,
  Le~Scao, Gugger, Drame, Lhoest, and Rush}]{wolf-etal-2020-transformers}
Thomas Wolf, Lysandre Debut, Victor Sanh, Julien Chaumond, Clement Delangue,
  Anthony Moi, Pierric Cistac, Tim Rault, Remi Louf, Morgan Funtowicz, Joe
  Davison, Sam Shleifer, Patrick von Platen, Clara Ma, Yacine Jernite, Julien
  Plu, Canwen Xu, Teven Le~Scao, Sylvain Gugger, Mariama Drame, Quentin Lhoest,
  and Alexander Rush. 2020.
\newblock \href {https://doi.org/10.18653/v1/2020.emnlp-demos.6} {Transformers:
  State-of-the-art natural language processing}.
\newblock In \emph{Proceedings of the 2020 Conference on Empirical Methods in
  Natural Language Processing: System Demonstrations}, pages 38--45, Online.
  Association for Computational Linguistics.

\bibitem[{Zhuang et~al.(2021)Zhuang, Wayne, Ya, and
  Jun}]{zhuang-etal-2021-robustly}
Liu Zhuang, Lin Wayne, Shi Ya, and Zhao Jun. 2021.
\newblock \href {https://aclanthology.org/2021.ccl-1.108} {A robustly optimized
  {BERT} pre-training approach with post-training}.
\newblock In \emph{Proceedings of the 20th Chinese National Conference on
  Computational Linguistics}, pages 1218--1227, Huhhot, China. Chinese
  Information Processing Society of China.

\end{thebibliography}
\bibliographystyle{acl_natbib}

\end{document}